\documentclass[runningheads]{llncs}
\usepackage[T1]{fontenc}
\usepackage{graphicx}
\usepackage{booktabs}
\usepackage{appendix}
\usepackage{float}
\usepackage{xcolor} 
\usepackage{CJKutf8}
\usepackage{url}
\usepackage{tabularx}
\usepackage{amsmath}
\usepackage{amssymb}
\usepackage{subcaption} 
\usepackage{multirow}

\usepackage[ruled]{algorithm2e}
\usepackage[misc]{ifsym}
\newcommand{\corr}{(\Letter)}
\usepackage{mwe}

\begin{document}

\title{TCMBench: A Comprehensive Benchmark for Evaluating Large Language Models in Traditional Chinese Medicine}

\titlerunning{TCMBench}


\author{Wenjing Yue\inst{1,2} \and
Xiaoling Wang\inst{1}\corr
Wei Zhu\inst{1} \and
Ming Guan\inst{1} \and
Huanran Zheng\inst{1} \and
Pengfei Wang\inst{1} \and
Changzhi Sun\inst{3} \and
Xin Ma\inst{4}
}



\institute{Shanghai Institute of AI for Education, East China Normal University, Shanghai, China \\ 
\and
School of Computer Science and Technology , East China Normal University, Shanghai, China \\ 
School of Pharmacy, East China Normal University, Shanghai, China \\
Shanghai Skin Disease Hospital, School of Medicine, Tongji University, Shanghai, China 
\email{\{wjyue, wzhu, hrzheng, pfwang\}@stu.ecnu.edu.cn} \email{xlwang@stu.ecnu.edu.cn} 
\email{czsun.cs@gmail.com}
\email{nicole.ma@me.com}
}

\maketitle              

\begin{abstract}
Large language models (LLMs) have performed remarkably well in various natural language processing tasks by benchmarking, including in the Western medical domain. However, the professional evaluation benchmarks for LLMs have yet to be covered in the traditional Chinese medicine(TCM) domain, which has a profound history and vast influence. To address this research gap, we introduce TCM-Bench, an comprehensive benchmark for evaluating LLM performance in TCM. It comprises the TCM-ED dataset, consisting of 5,473 questions sourced from the TCM Licensing Exam (TCMLE), including 1,300 questions with authoritative analysis. It covers the core components of TCMLE, including TCM basis and clinical practice. To evaluate LLMs beyond accuracy of question answering, we propose TCMScore, a metric tailored for evaluating the quality of answers generated by LLMs for TCM related questions. It comprehensively considers the consistency of TCM semantics and knowledge. After conducting comprehensive experimental analyses from diverse perspectives, we can obtain the following findings: (1) The unsatisfactory performance of LLMs on this benchmark underscores their significant room for improvement in TCM. (2) Introducing domain knowledge can enhance LLMs' performance. However, for in-domain models like ZhongJing-TCM, the quality of generated analysis text has decreased, and we hypothesize that their fine-tuning process affects the basic LLM capabilities. (3) Traditional metrics for text generation quality like Rouge and BertScore are susceptible to text length and surface semantic ambiguity, while domain-specific metrics such as TCMScore can further supplement and explain their evaluation results. These findings highlight the capabilities and limitations of LLMs in the TCM and aim to provide a more profound assistance to medical research. 
\keywords{Benchmark \and Traditional Chinese Medicine \and  Large Language Model \and Healthcare and Medicine.}
\end{abstract}

\section{Introduction}
Recently, Large language models (LLMs) have been demonstrated to lead performance in enhancing the accuracy of natural language understanding and text generation quality. Emerging studies have designed various LLMs like ChatMed\footnote{https://github.com/michael-wzhu/ChatMed}, HuaTuo~\cite{wang2023huatuo}, and ZhongJing-TCM~\footnote{https://github.com/pariskang/CMLM-ZhongJing}, highlights the growing demand for LLMs in the medical domain. Therefore, the standardized medical benchmark is essential for effectively developing and applying LLMs in medicine, providing reliable and authoritative assessments.

\begin{figure}[t]
    \centering
    \includegraphics[height=0.14\textwidth, width=0.72\textwidth]{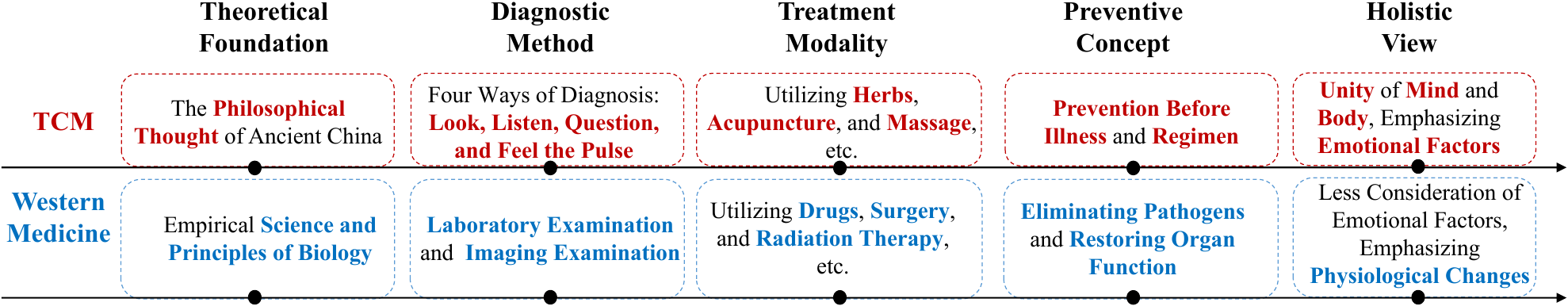}
    \caption{The difference between TCM and Western Medicine.}
    \label{fig:difference}
\end{figure}

Popular medical benchmarks mainly focus on Western medicine, such as MedMCQA~\cite{Medmcqa-2022} and MultiMedQA~\cite{MultiMedQA-2023}. Among this, MultiMedQA combines new online medical queries to alleviate data contamination issues in evaluations based on publicly easily accessible data sources~\cite{medbench-linlin2023}. Nonetheless, there are significant differences among medical systems, including variations in clinical standards, procedures, and languages~\cite{medbench-linlin2023}, particularly evident in the differences between Traditional Chinese Medicine(TCM) and Western Medicine. 
TCM has a long and rich history, making profound contributions to healthcare \cite{1-tcm}. Unlike Western evidence-based medicine, TCM emphasizes the clinical experience of physicians\cite{28-zhang2021computational}. Moreover, significant differences exist between their regarding theoretical foundations, diagnostic methods, treatment modalities, preventive concepts, and holistic views, as illustrated in Figure~\ref{fig:difference}. These differences highlight the unique aspects of TCM diagnosis, treatment, and knowledge in the medical field. 
While some benchmarking has been considered to evaluate the Chinese medical domain~\cite{medbench-linlin2023,CMExam-liu2024}, they also primarily evaluate modern Chinese medicine knowledge based on Western medicine principles. Therefore, directly applying existing Western medical benchmarks to assess TCM may not comprehensively evaluate the potential and practical utility of LLMs in this domain. Recently, the TCM LLM, such as ZhongJing-TCM, relies solely on physicians' subjective evaluation model performance, which consumes valuable time and leads to low efficiency. It highlights the urgent need for a standardized benchmark in TCM to provide objective and reliable evaluations of LLMs' performance.

\begin{figure}[t]
    \centering
    \includegraphics[height=0.4\textwidth, width=0.6\textwidth]{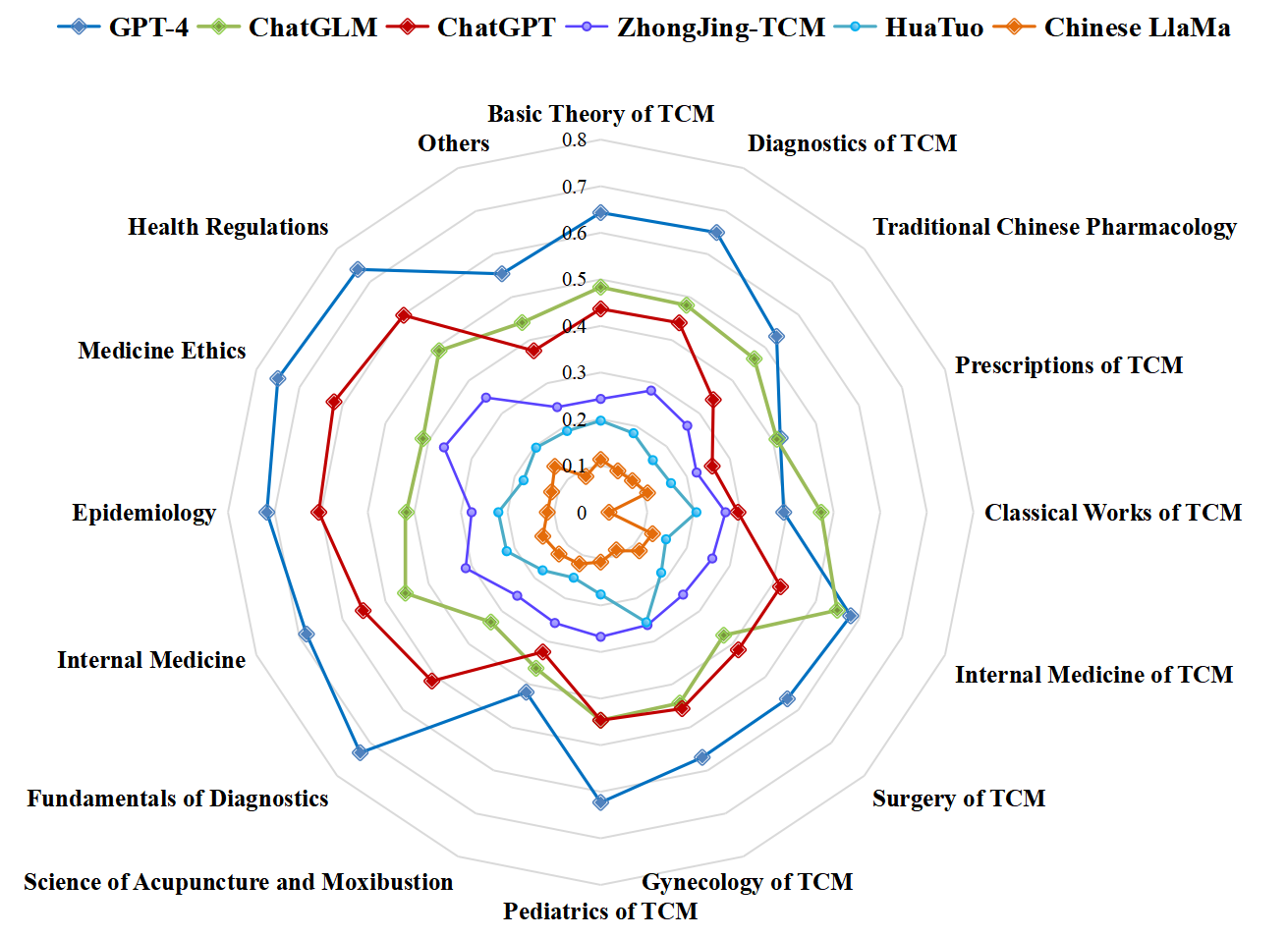}
    \caption{The performance of different LLMs on different branches of TCM-Bench.}
    \label{fig:kp_intro}
\end{figure}

To address these gaps and accommodate the unique characteristics of TCM knowledge, we introduce a new comprehensive benchmark, \textbf{TCMBench}, to supplement prior medical benchmarks. It is sourced from the TCM Licensing Exam (TCMLE), which is specially tailored for the TCM domain. To prevent data contamination, we construct a large-scale TCM evaluation dataset \textbf{TCM-ED} using actual TCMLE practice questions. It comprises 5,473 question-answer(Q\&A) pairs, with 1,300 data pairs with standard analysis, ensuring the reliability of the data quality. We manually select questions from the original dataset to confirm comprehensive coverage of all TCM branches, ensuring a broad scope within the topic of TCM-ED. The accuracy of LLMs in all TCM branches is shown in Figure\ref{fig:kp_intro}. Significantly, due to the unique terminology of TCM, exemplified by phrases like `catching a cold from the chilly wind' and `external attack of wind cold,' which convey the same meaning despite having low word matching. Thus, using traditional text generation metrics based solely on word matching \cite{CMExam-liu2024,medbench-linlin2023} or Chinese semantic representation models to evaluate the semantic consistency between two texts in TCM may not be appropriate. Based on these findings, we introduce an automatic metric called TCMScore to evaluate the consistency of TCM semantics and knowledge. It combines the matching of TCM terms and the semantic consistency between the generated and standard analyses. 

We conduct extensive experiments, combining various metrics, to provide a detailed analysis from different perspectives on evaluating LLMs' ability to understand, analyze, and apply knowledge in TCM.
The main findings of this benchmark are as follows:

\begin{itemize}
    \item The current performance of LLMs on this benchmark is unsatisfactory, indicating considerable room for improving their application in TCM. But general LLMs with hundreds of billions of parameters demonstrate potential for better application in TCM.
    \item Figure~\ref{fig:kp_intro} shows that general LLMs without specialized tuning are biased toward Western medicine. However, infusing professional TCM knowledge and related linguistic-cultural corpus can significantly improve LLMs' comprehension of context in TCM.
    \item From the express quality and human evaluation, fine-tuning LLMs with domain knowledge in TCM weakens their fundamental abilities in logical reasoning, knowledge analysis, and semantic expression. Thus, preserving these core abilities during pre-training is crucial.
    \item The expression quality metrics that rely on word matching or semantic similarity are easily affected by factors such as text length and surface semantic ambiguity. The TCMScore introduced in our work effectively addresses this limitation and can better supplement and explain the performance of LLMs in TCM semantic and knowledge consistency under the above metrics.
\end{itemize}

In conclusion, we propose an comprehensive benchmark aligned with the requirements of TCM, aiming to showcase LLMs' capabilities and limitations in the TCM domain and improve further developments in medical research. 

\section{Related Work}
As LLMs advance rapidly, benchmarking is crucial for driving progress in natural language processing (NLP), particularly in professional domains like medicine. Various medical Q\&A benchmarks tailored to different regions and medical systems have been instrumental in evaluating the efficacy of LLMs. For example, Kuang et al.~\cite{kung2023performance} utilize the United States Medical Licensing Examination(USMLE) questions to evaluate ChatGPT, while MedMCQA~\cite{Medmcqa-2022} develops a benchmark using Indian medical data. CBLUE~\cite{zhang2021cblue} and PromptCBLUE~\cite{promptcblue-zhu2023} focus on evaluation based on Chinese bio-medical information. Additionally, MultiMedQA~\cite{MultiMedQA-2023} combines six existing medical benchmark datasets, like MedQA~\cite{MedQA-jin2021} and MedMCQA, to mitigate data contamination through new online data sets. 
It also conducts a human evaluation through instruction fine-tuning based on the limited number of doctor-labeled samples. 
However, existing benchmarks focus on Western medical systems and lack TCM content. TCM, recognized by the World Health Organization as an effective complementary and alternative medicine system, differs significantly from Western medicine in its theoretical structure, diagnosis, and treatment standards~\cite{28-zhang2021computational}. Therefore,  Western medicine-based benchmarks cannot adequately evaluate the performance of LLMs in TCM.
Although benchmarks like CMExam~\cite{CMExam-liu2024} and MedBench~\cite{medbench-linlin2023} have been proposed for CNMLE, they also focus on modern Chinese medicine questions based on Western medicine principles. Moreover, they only offer rough statistics on TCM and briefly use metrics like Rouge for analysis. However, due to the unique term expression characteristics of TCM, these benchmarks cannot fully evaluate LLMs' performance in answering TCM questions. There remains a lack of comprehensive benchmarks in the academic community for evaluating LLMs in the TCM domain. In the evaluation of ZhongJing-TCM, only subjective evaluation by doctors was used to evaluate model performance. However, manual evaluation is time-consuming and labor-intensive, making it difficult to achieve large-scale applications. Overall, it is urgent to develop objective and systematic LLM evaluation benchmarks that adapt to the characteristics of TCM to fill the gap in evaluation standards in this domain.

\begin{figure}[t]
    \centering
    \includegraphics[height=0.47\textwidth, width=0.88\textwidth]{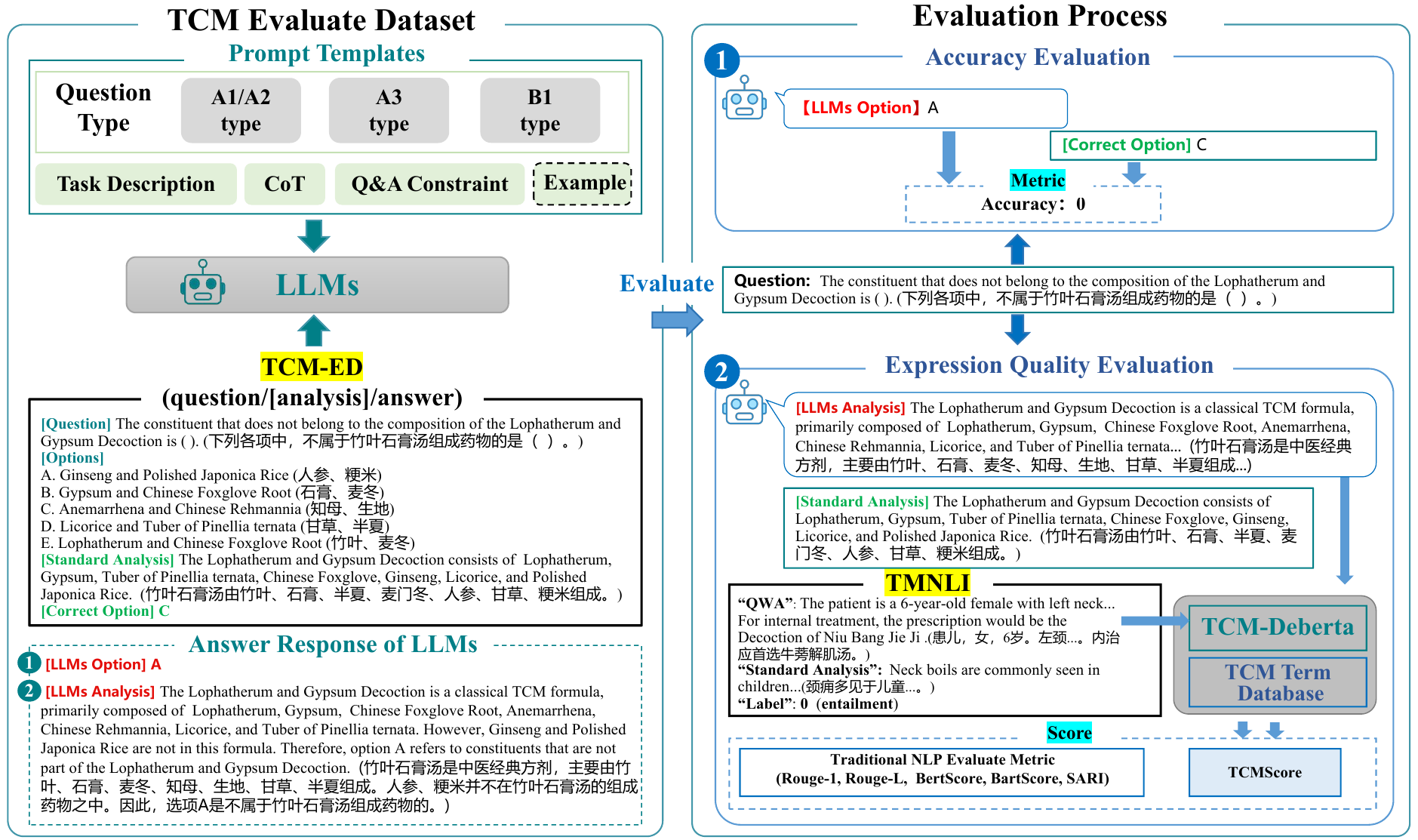}
    \caption{The overview of TCM-Bench. It consists of two parts: (1) On the left is the construction process of the \colorbox{yellow}{TCM-ED} dataset. (2) On the right is the evaluation process of TCM-Bench. The bottom section showcases the \colorbox{yellow}{TMNLI} dataset and the TCM-Deberta model, as well as the \colorbox[RGB]{235,240,255}{TCMScore} metric.
    }
    \label{fig:evaluate_process}
\end{figure}

\section{The Proposed Benchmark}
\subsection{Overview}
We propose an comprehensive benchmark, TCMBench, for evaluating the effectiveness of LLMs in TCM, as depicted in Figure~\ref{fig:evaluate_process}. It includes an evaluation dataset, TCM-ED, comprising 5,473 actual practice questions from TCMLE, which reflect the fundamental medical knowledge and reasoning logic required to obtain a TCM license in China. To create an automatic evaluation metric aligned with expert cognition, TCMScore, we first collect 9,788 recent real questions with analysis from TCMLE to build the first TCM natural language inference (NLI) dataset, TMNLI. We then introduce TCM-Deberta, a more stable semantic inference model, to effectively evaluate TCM semantic consistency. Additionally, we incorporate a task to evaluate TCM terminology matching in calculating TCMScore to reveal the knowledge consistency. In the end, we utilize multiple metrics to evaluate the ability and quality of LLMs to express TCM knowledge. 



\begin{figure}[t]
  \begin{minipage}[h]{0.4\textwidth} 
  \centering
    \includegraphics[height=0.6\textwidth, width=\textwidth]{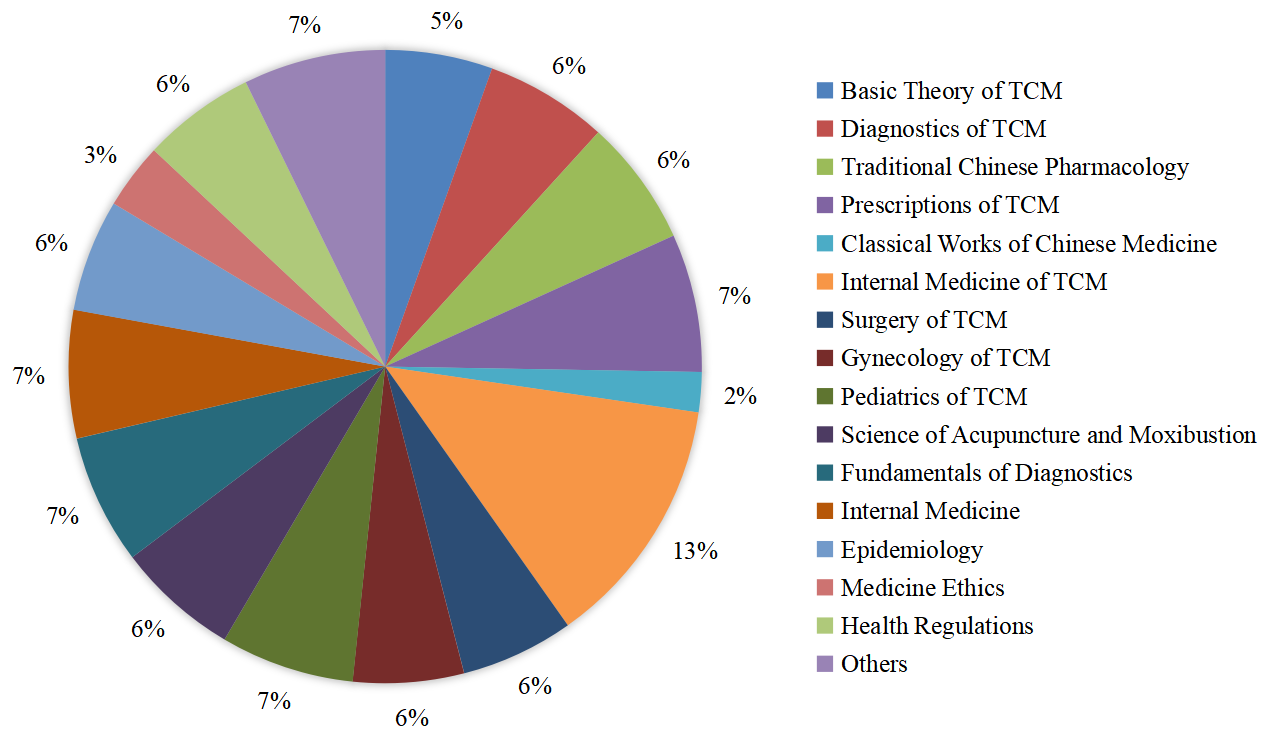}
    \captionof{figure}{Branches of TCM \\ in TCM-ED.}
    \label{fig:benches_classification}
  \end{minipage} \hfill
  \begin{minipage}[h]{0.6\textwidth} 
      \centering
        \captionof{table}{The statistical information of the TCM-ED (\textbf{upper}) and TMNLI(\textbf{lower}).}
        \resizebox{\linewidth}{!}{
         \begin{tabular}{c|l|c|c|c}
        \toprule
        \multirow{4}{*}{\textbf{TCM-ED}} & \textbf{Question Type} & \multicolumn{1}{l|}{\textbf{\#Qustions}}   & {\textbf{\#Sub-Question}} & \textbf{\#All} \\ \cmidrule{2-5} 
        & A1/A2  & 1,600  & 1,600 & \multirow{3}{*}{5473}  \\
        & A3 & 198& 642 &  \\
        & B1 & 1,481  & 3,231 &  \\ \hline \hline 
        \multirow{5}{*}{\textbf{TMNLI}}& \textbf{DataSet}   & \multicolumn{1}{l|}{\textbf{\#Entailment}} & \multicolumn{1}{l|}{\textbf{\#Contradiction}} & \multicolumn{1}{l}{\textbf{\#All}} \\ \cmidrule{2-5} 
           & Train  & 7,830  & \multicolumn{1}{c|}{17,751}   & 25,581 \\ 
           & Test   & 1,958  & \multicolumn{1}{c|}{1,958}& 3,916  \\ 
           & All& 9,788  & \multicolumn{1}{c|}{19,709}   & 29,497 \\ 
        \bottomrule
    \end{tabular}
        }
    \label{tab:TCM-ED-and-TMNLI}
  \end{minipage}
\end{figure}

\subsection{Construction and Statistics}
\subsubsection{TCM-ED}

The TCMLE assesses whether applicants possess the necessary professional knowledge and skills to practice as TCM physicians. Therefore, we collect 5,473 representative practice questions. Among them, the data we collect does not contain personal information but focuses on selecting data instances that can fully reflect and represent theoretical knowledge and practical skills in TCM. The multiple-choice questions in TCMLE are depicted on the left of Figure~\ref{fig:evaluate_process}, including three types:



\begin{itemize}
    \item  \textbf{The single-sentence best-choice questions(A1) and the case summary best-choice questions(A2) type}: It consists of a question stem and five options with correct one, as shown in Figure 1 of the appendix file.
    \item \textbf{The best choice questions for case group(A3) type}: The stem presents a patient-centered case, followed by multiple sub-questions, each offering five options with one correct answer. It primarily centers on clinical applications, as shown in Figure 2 of the appendix file.
    \item \textbf{The standard compatibility questions(B1) type}: Multiple sub-questions share the same five options, where each option may be chosen zero, one, or multiple times. There is one correct answer among the five options for each sub-question, as shown in Figure 3 of the appendix file.
\end{itemize}

Specifically, we manually screen the original practice question bank from TCMLE with the advice of experts to ensure that TCM-ED encompasses all question types and branches of TCM found in TCMLE. First, we cleaned the original data in PDF format. Then, we extract the questions, options, correct answers, and standard analysis using rule templates. Then, we convert the information into a structured JSON format. Subsequently, we randomly select 100 questions from the original question bank in each specific medical branch under each question type based on expert guidance. If a particular branch has fewer than 100 questions, all are selected. The detailed statistics of the TCM-ED are provided in the upper of Table~\ref{tab:TCM-ED-and-TMNLI}, indicating that each TCM branch in the A1/A2 type of questions comprises a complete 100 questions. Furthermore, Figure~\ref{fig:benches_classification} illustrates that the distribution of all TCM-ED questions across different branches is relatively balanced, ensuring that evaluation results are not biased due to the branch distribution. This ensures fairness and comprehensiveness in evaluation.

\subsubsection{TMNLI}
Due to the uniqueness of terminology in TCM, simple word matching or Chinese semantic similarity calculations cannot accurately measure semantic consistency in the TCM domain. In general, the NLI metrics are typically utilized to assess the fidelity of summaries. Previously work employ an entailment classifier trained on the MultiNLI(MNLI) dataset~\cite{MNLI} to determine if summaries are consistent with the context. However, since the MNLI dataset is in English, significant differences exist between it and TCM terminology. Therefore, we construct a TCM-specific NLI dataset dataset, TMNLI. We select 9,788 recent examination questions with standard analysis from TCMLE, covering three question types and all TCM branches depicted in Figure~\ref{fig:benches_classification}. Following the setting of the MNLI dataset, TMNLI consists of three parts: premise, hypothesis, and label. We leverage rule templates to combine the question and its correct answer into a claim called QWA, which serves as the premise. We then consider the standard parsing as a hypothesis and set the relation label between the two as entailment. Additionally, we utilize the BM25 algorithm to rank other analyses based on the similarity with QWA, randomly selecting up to three analyses from the top 20 to 100 rankings as hypotheses, labeled as contradictions. We aim to increase the difficulty in identifying the relationship between the premise and hypothesis. Consequently, we generated 29,497 NLI data. Inspired by previous work~\cite{semantic-uncertainty}, we consider that if there is stable semantic consistency between two sentences, it should be possible to derive the hypothesis from the premise and vice versa. Given the significant difference in length between QWA and the standard analyses in TMNLI, we permute the premises and hypotheses of half of the data pairs in TMNLI to eliminate bias caused by length discrepancies. The statistical information is detailed in the lower of Table~\ref{tab:TCM-ED-and-TMNLI}. 

\begin{table}[t]
\centering
 \caption{The statistical information of LLMs.}
  \label{tab:LLMs}
\resizebox{0.8\linewidth}{!}{
  \begin{tabular}{p{30mm}p{20mm}<{\centering}p{25mm}<{\centering}p{25mm}<{\centering}}
    \toprule
   Model & Open Source & Parameters & Domain \\
    \midrule
    GPT-4~\cite{openai-gpt4} & × & 175B+ & General \\
    ChatGPT~\cite{ChatGPT3-5} & × & 175B & General\\
    ChatGLM~\cite{glm-130B} & \checkmark & 130B & General\\
    Chinese LlaMa~\cite{chinese-llama-and-alpaca} & \checkmark & 7B & General\\
    HuaTuo\cite{wang2023huatuo} & \checkmark & 7B & Chinese Medicine\\
    ZhongJing-TCM\footnote{https://github.com/pariskang/CMLM-ZhongJing} & \checkmark & 7B & TCM \\
  \bottomrule
\end{tabular}
}
\end{table}

\section{Evaluation}
\subsection{Models}
To assess the medical abilities in the TCM domain, we utilize TCMBench to evaluate various LLMs across general and medical domains.
Specifically, we leverage the LLMs on an over 100 billion scale, such as the commercial (closed-source) GPT-4 and ChatGPT, as well as the open-source ChatGLM that supports Chinese. Additionally, we evaluate the general Chinese model Chinese-LlaMA and the Chinese medical-specific model HuaTuo that focuses on Western medicine, both fine-tuned from LlaMa-7B. Zhongjing-TCM is specialized in TCM gynecology Q\&A tasks. The statistical information is presented in Table~\ref{tab:LLMs}.

\subsection{Experimental Settings}
We conduct extensive experiments to evaluate the zero-shot performance of LLMs, ensuring their capability to respond in a multiple-choice format and provide corresponding analyses. Additionally, we partition the TCM-ED dataset based on medical branches and question types, conducting independent tests on each subset for comprehensive analysis. Depending on the question type, we design different prompt templates, including task descriptions, Chain-of-Through (CoT), and Q\&A constraints. The task description clarifies the question types that LLMs need to answer. The CoT guides LLMs in giving the options and providing corresponding analyses simultaneously, which can evaluate the ability of LLMs to understand and express TCM knowledge comprehensively. The specific content of the prompt template is shown in Section B of the appendix file in additional materials. Particularly, due to the shared content among several questions of types A3 and B1, strong logical coherence exists between the questions. To evaluate the logical reasoning ability of LLMs in TCM, we adopt a multi-turn dialogue format, using answers from preceding questions as historical context for subsequent dialogues. Moreover, we observe that the A3 type of questions closely resemble real-world clinical diagnosis and treatment processes, yet requiring LLMs to answer in a fixed format poses considerable difficulty. Hence, in such questions, we introduce the few-shot to incorporate an A3-type answer example at the beginning of the question as a prompt for the answering format.

\begin{table}[t]
\centering
  \caption{The accuracy of different NLI models on two testing datasets. }
  \label{tab:nli_result}
  \begin{tabular}{l|c|c}
    \toprule
    \textbf{Model} & \textbf{TMNLI(Test)}  & \textbf{TCM-ED(Test)}\\
    \midrule
    DeBERTa-v3-base & 33.2\%& 0.0\%\\
    DeBERTa-v3-base-mnli & 41.5\%& 4.08\%\\
    TCM-Deberta & \textbf{96.48\%}& \textbf{95.38\%}\\
  \bottomrule
\end{tabular}
\end{table}

\subsection{Evaluation Metrics}
We evaluate both general and medical LLMs using TCM-Bench, following the evaluation process depicted on the right of Figure~\ref{fig:evaluate_process}, which comprises two key steps. Firstly, we employ accuracy as the evaluation metric to automatically compare the options generated by LLMs with the correct options, evaluating their understanding and application ability for TCM knowledge. Secondly, we choose 1,300 questions with standard analysis from TCM-ED to automatically evaluate the quality of LLMs in expressing knowledge in TCM. We employ traditional metrics of text generation tasks, including word matching methods like ROUGE~\cite{lin2004rouge} and SARI~\cite{SARI}, as well as deep learning methods like BertScore~\cite{zhang2019bertscore} and BartScore~\cite{yuan2021bartscore}, to compare the semantic similarity between the analysis generated by LLMs and the standard analysis. Additionally, we introduce the expert-level supplement metric TCMScore, which reflects TCM semantics and knowledge consistency. The two evaluation processes complement each other, providing a more comprehensive perspective on evaluating the medical performance of LLMs in TCM.

Now, we introduce TCMScore. Firstly, we fine-tune the NLI model, DeBERTa-v3-base-mnli~\footnote{https://huggingface.co/MoritzLaurer/DeBERTa-v3-base-mnli}, to create TCM-Deberta tailored for inferring TCM semantic consistency between two sentences. To further illustrate its effectiveness, we evaluate the inference accuracy of different NLI models on the TMNLI test set. Moreover, we evaluate the model accuracy of predicting the relationship between standard analysis and QWA in TCM-ED, whose data differ from TMNLI. Among this, we employ a more stable method in which set analysis and QWA are both the premise and hypothesis. When the two inference results are the entailment, we consider a stable semantic consistency between them. The inference results presented in Table~\ref{tab:nli_result} demonstrates that TCM-Derberta achieves stable and high accuracy on two testing datasets.

In addition, to evaluate the TCM knowledge consistency between texts, we design a metric,Term F1 Score($F1^*$), which quantitatively measures the matching score of TCM terms between two text. The core idea of the $F1^*$ is to comprehensively consider the redundancy(i.e., precision), matching degree (i.e., recall), and term diversity of TCM terminologies. The $F1^*$ is as follows.
\begin{small}
\begin{equation}
\begin{split}
    Precision &= \dfrac{|M(T(S_1), T(S_2))|}{T(S_1)}, \\
    Recall &= \dfrac{|M(T(S_1), T(S_2))|}{T(S_2)}, \\
    Diversity &= \dfrac{|M(T(S_1), D)|}{T(S_1)}, \\
    F1^* = 3 \times &\dfrac{Precision \times Recall \times Diversity}{Precision + Recall + Diversity}.
\end{split}
\label{eq:termF1}
\end{equation}
\end{small}
Guided by experts, we standardize 61,987 TCM terminologies from official publications on TCM diagnosis and treatment, TCM disease and syndrome codes, and TCM-KB\cite{wangxinyu_topic_model} to create the TCM terminology database, $D$.
$T(S)=Counter(S) \cap Counter(D)$ represents the set of TCM terminologies and their quantities in sentence $S$, and $M(T_1, T_2) = T_1 \cap T_2$ denotes the matching quantities of TCM terminologies between two sets. 

\begin{algorithm}[t]
 \caption{The calculation process of TCMScore.}
 \label{alg:1}
 \KwIn{The LLMs generated sentences $S$,
       The standard analysis sentence $A$}
  \KwOut{$TCMScore$}
  \For{$s_{a} \in A$}
  { 
    \For{$s_{LLM} \in S$}
      {
        $F1^* \leftarrow$ calculate Term F1 score between $s_{LLM}$ and $s_{a}$; \\
        $nli_{score} \leftarrow$ (TCM-Deberta($s_{LLM}$, $s_{a}$) $+$ TCM-Deberta($s_{a}$, $s_{LLM}$)) / 2; \\
        add $F1^*$ to a list $F1^*_{LLM}$, and
        add $NLI_{score}$ to a list $Senmentic_{LLM}$;\\
       }    
    \For{$i \gets 1$ \KwTo length($F1^*_{LLM}$)}
    { 
      $W_{senmentic} \leftarrow F1^*_{LLMs}[i] / \sum{F1^*_{LLM}} \times Senmentic_{LLM}[i] $;\\
      add $W_{senmentic}$ to a list $L_{TCMScore}$;\\
    }   
  }
  $w_{length} = e^{1-log(max(|S|, |A|) / log(min(|S|, |A|)}$ ; \\
  $TCMScore \leftarrow \sum{L_{TCMScore}} / length(L_{TCMScore}) \times w_{length} $;\\
\Return $TCMScore$ \\
\end{algorithm}

Finally, we combine $F1^*$ and the TCM-Deberta model to construct TCMScore. Its essence lies in focusing more on the semantic consistency of the LLMs generated sentence when its TCM terms match numerous with the standard analysis. Conversely, if the term matching is low, even if the semantics of the sentence are similar to the analysis, it appears relatively unimportant. Since the evaluated text is lengthy, we employ a sentence-by-sentence analysis method. We calculate the $F1^*$ score between each standard analysis sentence and LLMs-generated sentence to measure the knowledge matching degree. This score is then normalized as the importance weight when evaluating the semantic consistency of the sentence in LLMs-generated response. Next, we use the TCM-Deberta to calculate the semantic consistency score between each pair of sentences, which is then multiplied by weight to obtain the weighted semantic consistency score. Finally, we summarize the scores of each sentence to determine the difference between the analysis generated by LLMs and the standard analysis, resulting in overall TCM semantic and knowledge consistency scores. Moreover, we introduce a length penalty term $w_{length}$ to balance the impact of text length differences. It imposes a more significant penalty on short text than on long text. The calculation process is outlined in Algorithm~\ref{alg:1}.

\subsection{Main Results}
\subsubsection{Analysis of LLMs Accuracy.}
From Table~\ref{tab:result}, we comprehensively analyze the accuracy for different LLMs in TCMLE. The main findings are as follows.

\begin{small}
\begin{table}[t]
\centering
  \caption{The accuracy on LLMs in three question types of TCM-Bench. The best performing model is bold, while the strongest models are underlined.}
  \label{tab:result}
  \begin{tabular}{l|c|c|c|c|c}
    \toprule
    \textbf{LLMs} & \textbf{A1/A2} & \textbf{A3 (zero-shot)}& \textbf{A3(few-shot)}&\textbf{B1} & \textbf{Total}\\
    \midrule
     Chinese LlaMa & 0.0969& 0.1075&0.1620& 0.1151&0.1089\\
    HuaTuo & 0.1944& 0.1981&0.1402& 0.1876&0.184\\
    ZhongJing-TCM & 0.3537 & 0.3364&0.3178& 0.2182&0.2695\\ \midrule
    ChatGLM & 0.3613& 0.4595&\underline{0.6168}& \underline{0.4568}& \underline{0.4477}\\
    ChatGPT & \underline{0.4510}& \underline{0.4657}&0.4782& 0.4444& 0.4398\\
    GPT-4 & \textbf{0.5819}& \textbf{0.6231}&\textbf{0.6277}& \textbf{0.6011}& \textbf{0.5986}\\
  \bottomrule
\end{tabular}
   
\end{table}
\end{small}
\textbf{(1) None of the tested LLMs passed the TCMLE.} A notable observation is that accuracy improves with increasing model parameters. GPT-4 consistently outperforms other LLMs despite some being trained on extensive Chinese or medical corpora. Even so, the total accuracy of GPT-4 does not exceed 60\%, which is the minimum requirement for passing TCMLE. This also indicates that there is still significant room for improvement in the medical performance of LLMs in the TCM domain.



\textbf{(2) In the pre-training stage of LLMs, incorporating domain-specific knowledge becomes more crucial in the same magnitude order of parameters.} Despite having over 100 billion parameters, the overall accuracy of ChatGPT is lower than ChatGLM.  It is due to ChatGLM using a more extensive Chinese corpus during the pre-training stage that its ability to understand Chinese-based TCM questions is enhanced. However, the slight difference in overall accuracy between the two models highlights the gap between general Chinese and TCM semantics. In the 7 billion parameter level LLMs, incorporating professional knowledge of medicine, especially TCM, during pre-training can significantly improve model performance. For example, ZhongJing-TCM achieves an accuracy of 35.37\% on the A1/A2 type of questions, which is only 2\% lower than ChatGPT despite a 25-fold difference in the parameter numbers. The comparison powerfully illustrates that merely increasing the parameter numbers is not optimal when dealing with specific domains like TCM, which has profound cultural backgrounds and professional terminology. Instead, carefully designing and integrating high-quality professional TCM data during the pre-training stage is an effective way to enhance the performance of LLMs in TCM applications.

\textbf{(3) Adding examples to prompts can enhance the ability of LLMs to handle complex logical reasoning.} We can observe the unsatisfactory zero-shot performance on ChatGLM, ChatGPT, and Chinese LlaMa, as shown in Table~\ref{tab:result}. However, the performance of these models significantly improved once examples are introduced, with ChatGLM showing a performance increase of 34.23\%. This demonstrates that designing compelling TCM examples can enhance LLMs' understanding of complex reasoning logic in the TCM field.


\textbf{(4) Adding examples in prompts to guide LLMs through complex logical reasoning may not necessarily be effective.} Incorporating domain knowledge may damage the model's original logical reasoning ability in the fine-tuning process of LLMs. Their performance decreases when HuaTuo and Zhongjing-TCM engage in few-shot learning through examples. The possible reason is that adding examples would create a longer prompt, which exceeds the capability of LLMs to handle long texts. Therefore, in the future, while enhancing the domain adaptability of LLMs, it is crucial to maintain and optimize their ability to handle complex and lengthy text logical tasks.
\begin{figure}[t]
\begin{minipage}[h]{0.5\textwidth} 
  \centering
    \includegraphics[height=0.45\textwidth, width=\textwidth]{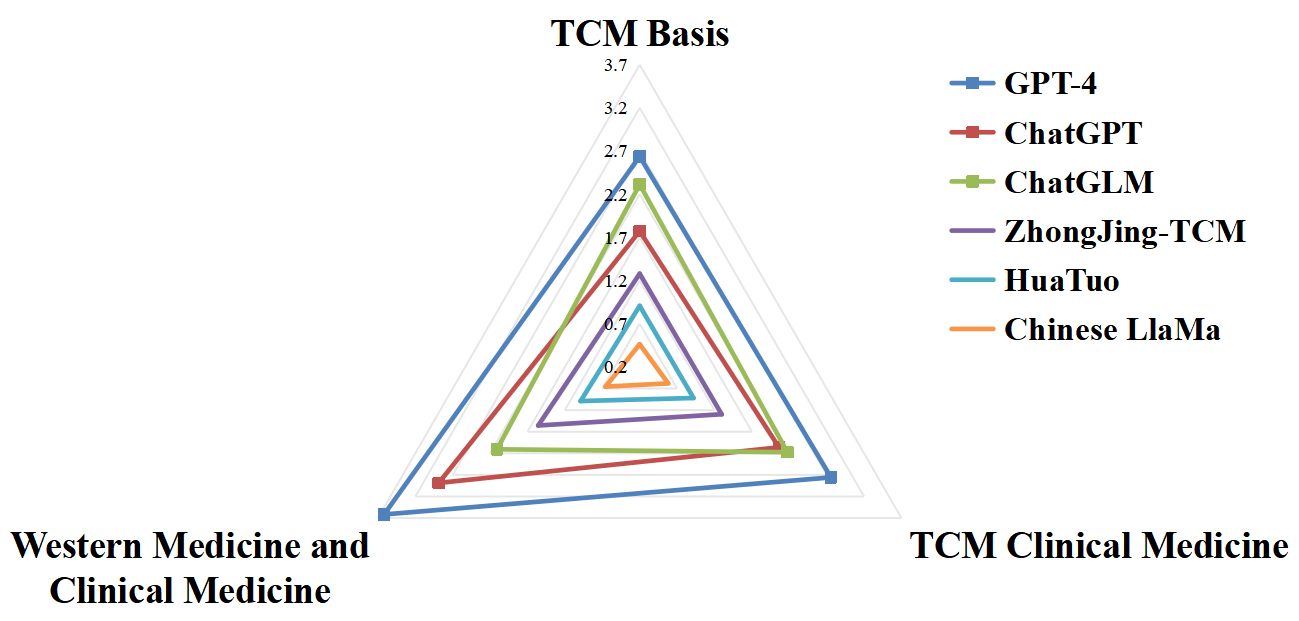}
    \caption{The total accuracy results on \\ category of TCMLE.}
    \label{fig:kp_result}
  \end{minipage} \hfill
\begin{minipage}[h]{0.5\textwidth} 
     \centering
        \captionof{table}{The comparative evaluation of GPT-4 and HuaTuo in answering to A1/A2 questions under various prompts.}
        \resizebox{\linewidth}{!}{
         \begin{tabular}{l|c|c|c}
            \toprule
            \textbf{Model} & \textbf{CoT}  & \textbf{No CoT} & \textbf{Answer 5 times} \\
            \midrule
            GPT-4 & 0.5819& 0.5475& 0.5475\\
            HuoTuo & 0.1944& 0.185& 0.1856\\
          \bottomrule
        \end{tabular}
        }
        \label{tab:cot}
\end{minipage}
\end{figure}

\begin{figure}[t]
    \centering
    \includegraphics[height=0.5\textwidth, width=0.97\textwidth]{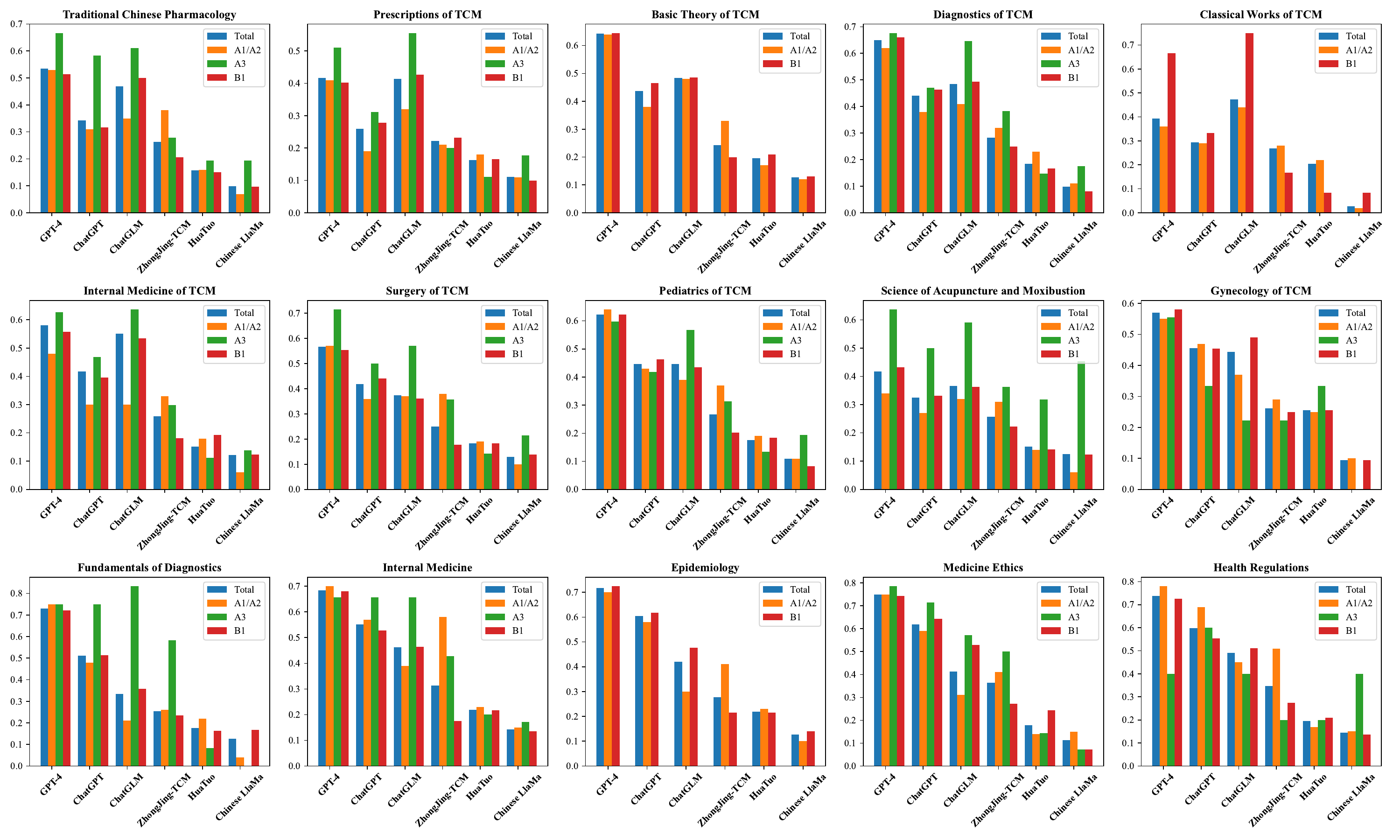}
    \caption{The accuracy results on different branches of TCM-Bench.}
    \label{fig:kp_result_combine}
\end{figure}

\textbf{(5) LLMs perform differently across various medical branches.} We further evaluate their accuracy in addressing questions within different medical domains. Firstly, based on the scope of the TCMLE exam, we categorize medical branches in TCM-Bench into three groups: TCM Basis, TCM Clinical Medicine, as well as Western Medicine and Clinical Medicine. The accuracy of different models in each category is illustrated in Figure~\ref{fig:kp_result}, with each category comprising five medical branches corresponding to subplots of a row in Figure~\ref{fig:kp_result_combine}. Figure~\ref{fig:kp_result_combine} details the model's accuracy across each question type. If LLMs perform well on A3 questions simulating clinical scenarios, they are better suited for complex clinical case analysis tasks. Additionally, the high accuracy of A3 and B1 questions through multiple rounds of dialogue indicates their effective understanding and correlation of various medical knowledge points. A notable finding is that GPT-4 and ChatGPT perform well in Western medicine. In contrast, ChatGLM excels in TCM Basis, especially in correlating and analyzing the knowledge of classic works of TCM. It highlights the importance of the Chinese corpus in comprehending theoretical TCM knowledge, but more professional knowledge is needed for clinical assistance. Although ZhongJing-TCM is trained based on the corpus generated from TCM gynecological medical records, it performs well in all branches, surpassing ChatGPT on the A1/A2 question of five branches, like Traditional Chinese Pharmacology. It also surpasses ChatGLM on four Western medical branches' A1/A2 questions. This demonstrates the model's performance in cross-domain knowledge transfer and comprehensive application.

\textbf{(6) Chain-of-Thought prompting and model stability.} 
In TCM-Bench, we set up CoT-based prompts for evaluation. In addition, we compare LLMs without using CoT prompts, and the accuracy is shown in Table~\ref{tab:cot}. After removing CoT, the overall performance of LLMs declined, confirming the importance of CoT cues in enhancing model comprehension of TCM knowledge. To evaluate the stability of LLMs without CoT, we take LLMs by answering each question 5 times and calculate the average score. The results showed that LLMs can still maintain high stability in this scenario.

\subsubsection{Analysis of expression quality of LLMs.}

We leverage three types of evaluation metrics: (a) word matching based methods like Rouge-1, Rouge-L, and SARI, (c) deep leaning based methods like BertScore, and BartScore, and (c) a hybrid method to professional term matching and deep learning, i.e., TCMScore to evaluate the LLMs' expression quality comprehensively. From Table~\ref{tab:express_result}, We find GPT-4 consistently demonstrates outstanding performance, but ZhongJing-TCM is the worst. Further findings are as follows.

\begin{figure}[t]
\begin{minipage}[c]{0.49\textwidth} 
  \centering
    \includegraphics[height=0.5\textwidth, width=0.9\textwidth]{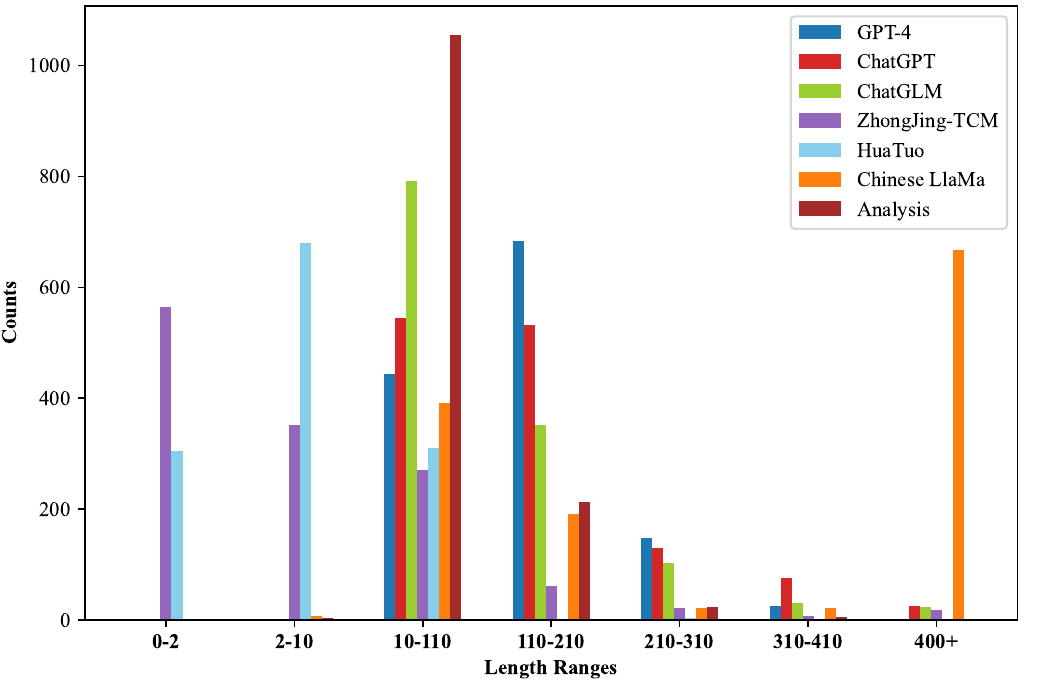}
    \caption{LLMs generated text length information statistics.}
    \label{fig:kp_result}
  \end{minipage} \hfill
\begin{minipage}[c]{0.45\textwidth} 
    \centering
    \includegraphics[height=0.5\textwidth, width=0.8\textwidth]{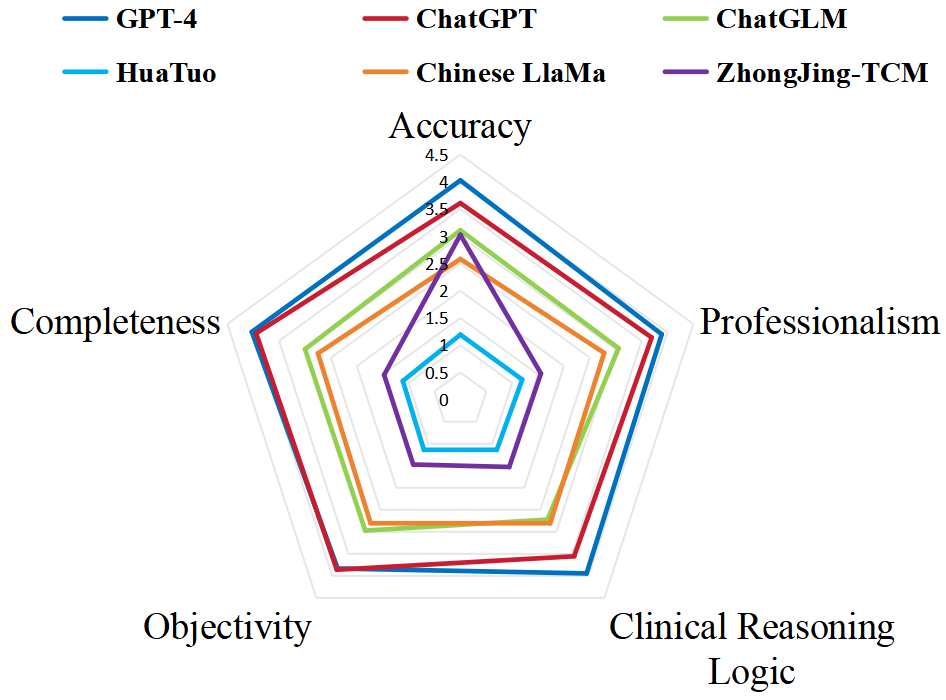}
    \caption{Human evaluation result.}
    \label{fig:response_length}
\end{minipage}
\end{figure}

\textbf{(1) Metrics based on word matching are affected by the length of the generated text.} Rouge favors LLMs with minimal differences in length compared to the standard analysis. Analyzing the Rouge score and the generated text length that depicted in Figure~\ref{fig:response_length}, ChatGPT and ChatGLM have an advantage due to their generated text length being closer to the standard analysis. Additionally, Rouge focuses on the recall metric, meaning that if the shorter text that LLMs generated, it may score higher. This explains why ZhongJing-TCM, generating less content yet maintaining higher accuracy, outperforms HuaTuo in Rouge score, while HuaTuo surpasses Chinese LlaMa. From the statistics in Figure 7 for the 0-2 to 2-10 length range, it is evident that HuaTuo and ShenNong-TCM struggle to generate analysis, emphasizing the need to maintain fundamental analysis and reasoning capabilities when fine-tuning domain knowledge again. We calculate the SARI score with retention rate due to without extra reference text. It considers word frequency to evaluate the information content of the generated text. For instance, GPT-4 excels in this metric. However, LLMs generating longer texts like Chinese LlaMa can achieve higher SARI scores despite potential errors due to containing more information. The above analysis confirms that relying solely on word matching to evaluate TCM text generation quality introduces bias due to text length.

\begin{table}[t]
  \caption{The expression quality of the response of LLMs on TCM-Bench. The best performing model is bold, while the strongest models are underlined.}
  \centering
  \label{tab:express_result}
  \resizebox{0.9\linewidth}{!}{
  \begin{tabular}{l|c|c|c|c|c|c}
    \toprule
    \textbf{LLMs} & \textbf{Rouge-1} & \textbf{Rouge-L}& \textbf{SARI} &  \textbf{BertScore}&\textbf{BartScore} & \textbf{TCMScore}\\
    \midrule
    ZhongJing-TCM & 3.41\% & 3.38\%&3.14\%& 48.01\%&-6.06&7.84 \%\\ 
    HuaTuo & 2.52\% & 2.48\%&2.95\%& 52.01\%&-5.45&8.8 \%\\
    Chinese LlaMa & 2.01\% & 1.93\%&9.99\%& 56.92\%&-4.4& 10.48\%\\ \midrule
     ChatGLM & \underline{4.85\%}&\underline{4.7\%}& 21.58\% &\underline{67.13\%}& -4.17&\underline{43.49}\%\\
    ChatGPT & \textbf{4.93\%} & \textbf{4.8\%}& \underline{23.93\%}& 66.95\% &\underline{-3.92} & 43.39\%\\
    GPT-4 & 4.8\% & 4.6\%& \textbf{25.34\%}&\textbf{67.58\%}& \textbf{-3.91}&\textbf{45.71\%} \\
  \bottomrule
\end{tabular}
}
\end{table}

\textbf{(2) Deep learning-based metrics like BertScore and BartScore can evaluate the semantic similarity of generated texts but do not directly evaluate or interpret the accuracy of expertise in the texts.} GPT-4 maintains its super in the two metrics, indicating high semantic similarity between its generated text and standard analysis. ChatGLM surpasses ChatGPT in BertScore, indicating a better ability to produce concise and focused responses. The analysis of the generated text length from Figure~\ref{fig:response_length} further supports this. Moreover, Chinese LlaMa outperforms HuaTuo and ZhongJing-TCM in both metrics, indicating its strength in generating diverse and complex semantic content. The texts it produces are often longer and offer more detailed analysis and expansion. However, it is essential to note that the length and semantic coherence of the generated text do not guarantee the absolute correctness of the content. In summary, evaluating text in a knowledge-intensive domain like TCM requires evaluating not just language fluency and semantic consistency but also specific metrics related to professional knowledge accuracy and domain adaptability. This ensures a comprehensive quality evaluation of the text.

\textbf{(3) Further, the findings will be supplemented by using TCMScore.} TCMScore serves as a comprehensive supplementary evaluation metric, which considers both semantic and knowledge consistency in TCM, along with the length of the generated text. With the introduction of TCMScore, we can draw the following conclusions to complement the findings above: 

\textbf{(a) GPT-4 can generate highly accurate content with rich TCM characteristics and expand knowledge boundaries. }Its significant advantages on TCMScore highlight its prowess. Despite potentially lower scores on Rouge, this underscores the advantage of GPT-4 in providing in-depth expansion and refinement while ensuring semantic consistency with the standard analysis. It offers a more comprehensive and detailed analysis related to TCM.

\textbf{(b) Wrong facts in the LLMs generated text diminish their advantage in semantic similarity.} ChatGPT and Chinese LlaMa can expand TCM knowledge, but the accuracy of their generated content still needs to be improved. While ChatGPT outperforms ChatGLM in most metrics after integrating TCMScore, its advantage diminishes due to the redundancy and errors in the generated content. Chinese LlaMa faces a similar issue, as evidenced by its reduced advantage in TCMScore despite excelling in BartScore and BertScore. This indicates a higher likelihood of incorrect information in its generated content to decrease its semantic advantage. Overall, these LLMs have room to enhance the accuracy of their generated content and reduce misinformation.

\textbf{(c) The correctness of text content does not equal the effective application of TCM knowledge.} For instance, while ZhongJing-TCM may provide accurate answers, it lacks essential skills like language fluency, knowledge coherence, and logical reasoning, resulting in poor performance on TCMScore. Therefore, to advance LLMs in TCM, enhancing their ability to apply knowledge is crucial while ensuring a deep understanding of domain-specific concepts.


\subsection{Human evaluation}

In addition to automatic evaluation, we invited a TCM expert and a medical doctoral student to conduct a manual evaluation of 18 questions related to TCM fundamentals and clinical knowledge to evaluate the performance of LLMs quantitatively. The evaluation dimensions include accuracy, professionalism, clinical reasoning logic, objectivity, and comprehensiveness, see Figure 8. For fairness, LLMs' names are anonymized during the evaluation process. The results revealed that GPT-4 excelled across all evaluation dimensions. ChatGPT showed slightly lower accuracy and clinical reasoning logic than GPT-4. Notably, Chinese LlaMa matched ChatGLM in objectivity and clinical reasoning logic despite its lower accuracy, which reflects that it can preserve its basic model's capabilities. ZhongJing-TCM exhibited a strong understanding of TCM knowledge on the accuracy dimension. However, it scored lower on other dimensions due to challenges in providing specific analyses (similar to HuaTuo).

\section{Conclusion}
In this paper, we introduce the TCMBench, a comprehensive benchmark for evaluating the performance of LLMs in TCM. The experiment shows that the performance of LLMs in this field is unsatisfactory. It also highlights the importance of maintaining the basic capabilities of LLMs while inducing domain expertise in their fine-tuning process. We also analyze the domain-specific metrics, like our TCMScore, which can further supplement and explain the evaluation results of traditional metrics for text generation. 

Furthermore, experiments reveal that some LLMs produce wrong information (i.e., hallucination phenomenon) when generating content. This area will be a focus of our future in-depth research aimed at developing effective methods to identify and quantify such issues. Considering the central role of clinical practice in TCM, we plan to expand the data source, which includes actual TCM case data, covering the entire TCM diagnosis and treatment process. Notably, we will concentrate on accurately evaluating if LLMs can follow the unique clinical logic of TCM, which is syndrome differentiation and treatment,  thereby enhancing and refining our benchmark.

%
%
%
\bibliographystyle{splncs04}
\bibliography{mybib}
%





\appendix
\section{The Question Type of TCM-EB}
\label{}
\begin{figure}[H]
    \centering
    \includegraphics[height=0.55\textwidth, width=0.9\textwidth]{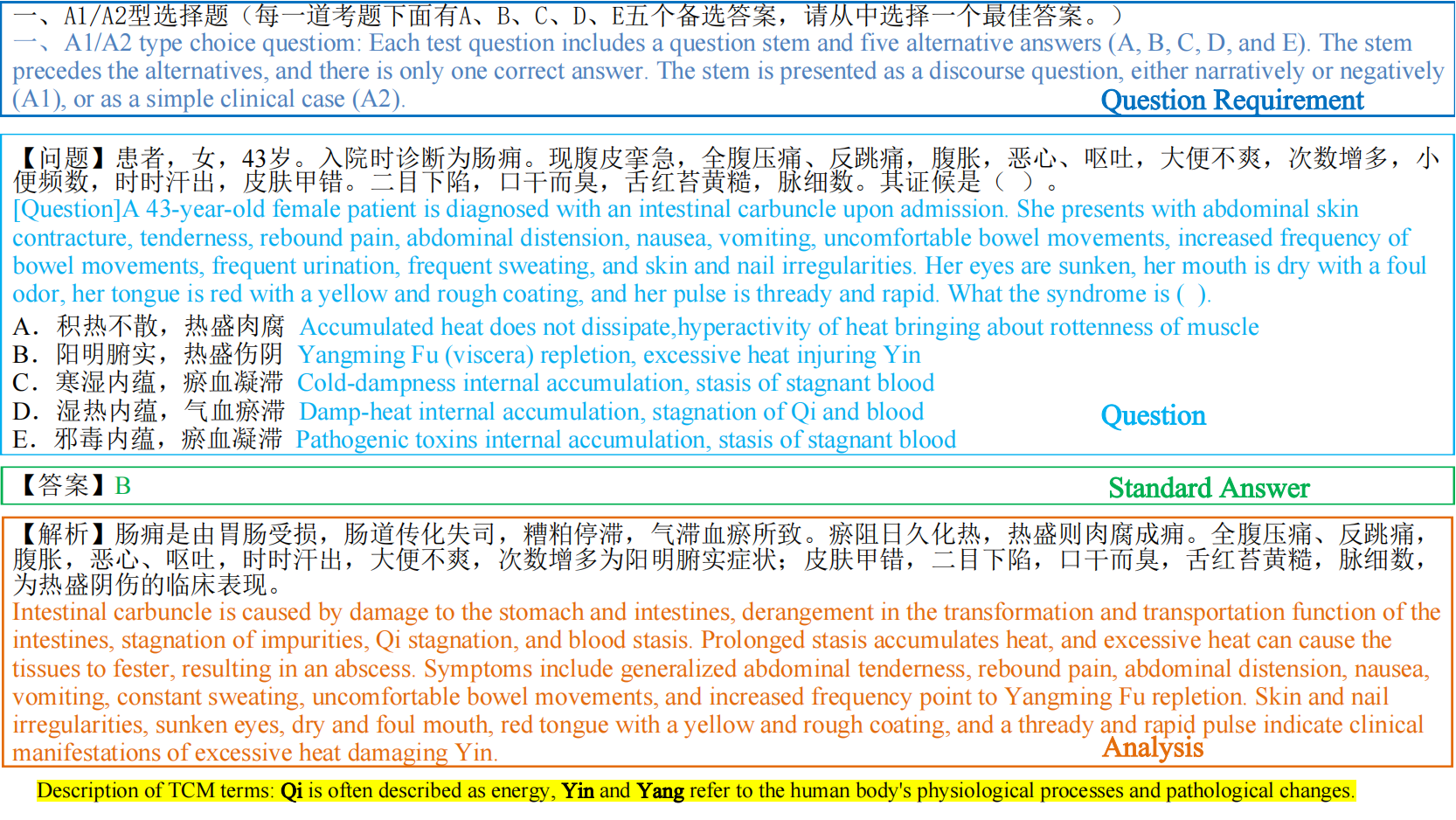}
    \caption{The example of A1/A2 type of questions. The \textcolor[rgb]{0.1,0.6,0.9}{question requirement} is indicated in dark blue text, \textcolor[rgb]{0.2,0.8,0.9}{the question along with the five options} is in light blue text, \textcolor[rgb]{0.0,0.8,0.5}{the standard answer} is in green text, and \textcolor{orange}{the standard analysis} is in orange text. The related \colorbox{yellow}{TCM terms} are explained in the yellow highlight.}
    \label{fig:A1-2}
\end{figure}

\begin{figure}[H]
    \centering
    \includegraphics[height=0.55\textwidth, width=0.8\textwidth]{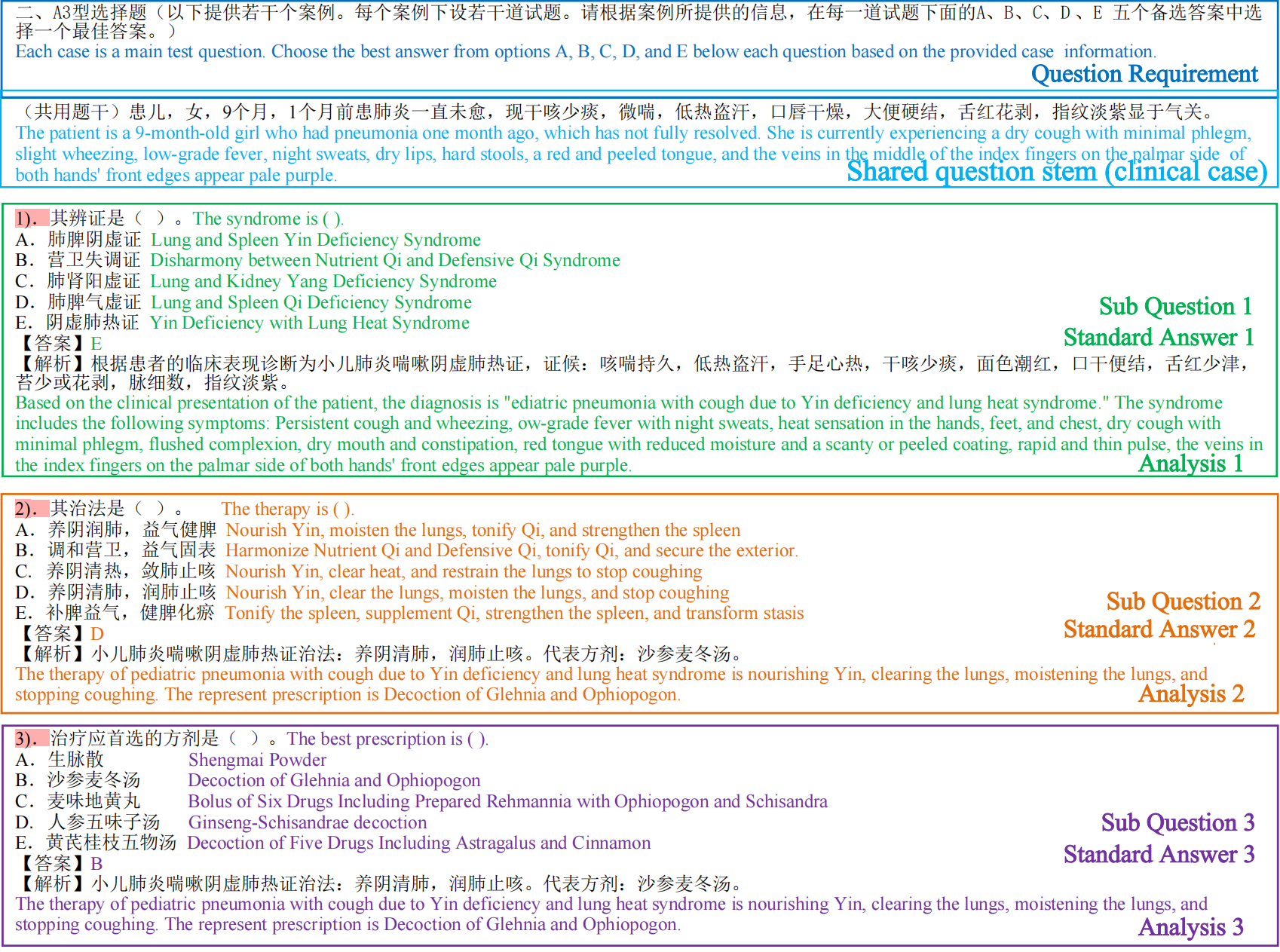}
    \caption{The example of A3 type of questions. The \textcolor[rgb]{0.1,0.6,0.9}{question requirement} is indicated in dark blue text, \textcolor[rgb]{0.2,0.8,0.9}{the patient-centered case} is in light blue text, \textcolor[rgb]{0.0,0.8,0.5}{the first sub question along with the five options, standard answer and analysis} is in green text, \textcolor{orange}{the second sub question} is in orange text, and \textcolor[rgb]{0.6,0.2,0.9}{the second sub question} is in purple text.}
    \label{fig:A3}
\end{figure}

\begin{figure}[H]
    \centering
    \includegraphics[height=0.4\textwidth, width=0.8\textwidth]{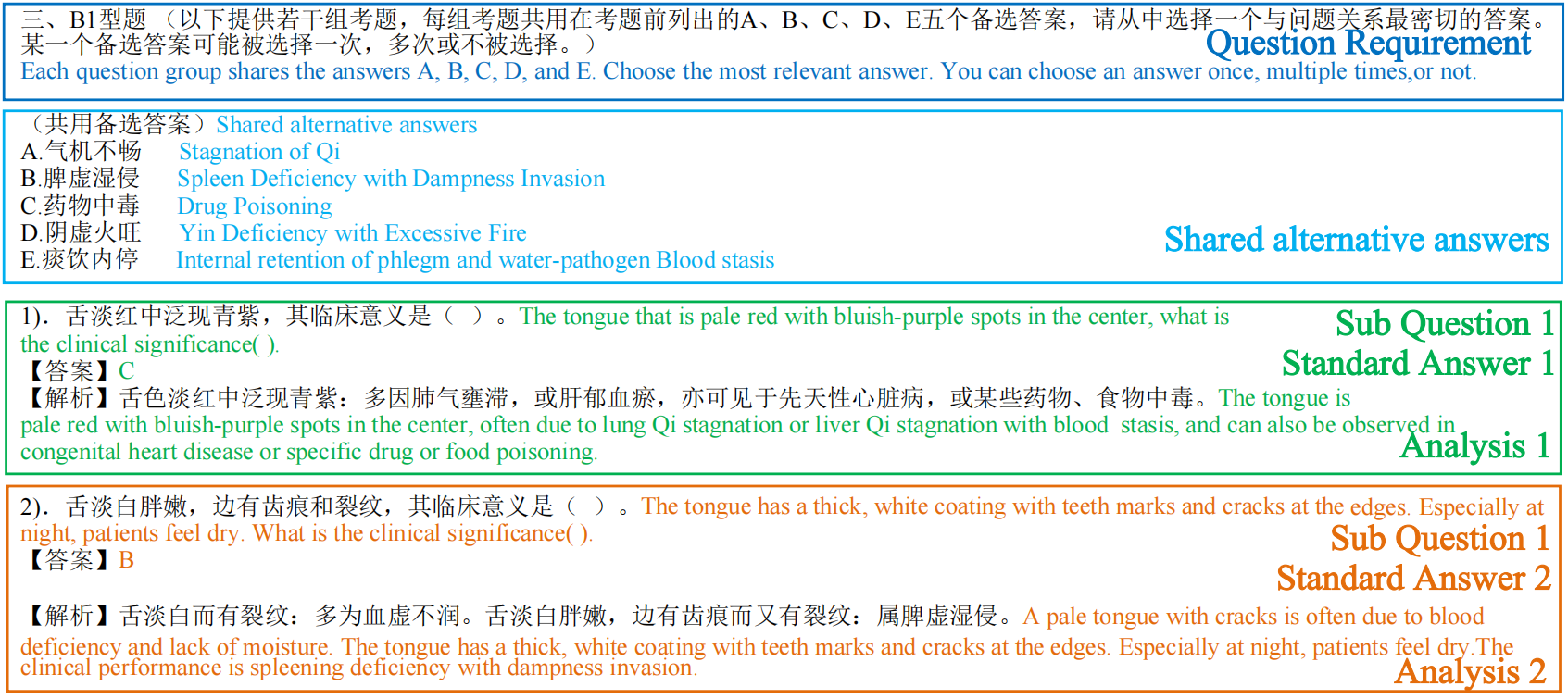}
    \caption{The example of B1 type of questions. The \textcolor[rgb]{0.1,0.6,0.9}{question requirement} is indicated in dark blue text, \textcolor[rgb]{0.2,0.8,0.9}{five options} is in light blue text, \textcolor[rgb]{0.0,0.8,0.5}{the first sub question along with the five options, standard answer and analysis} is in green text, and \textcolor{orange}{the second sub question} is in orange text.}
    \label{fig:B1}
\end{figure}

\section{Prompt and target output formats for Evaluation}
\label{prompt:evaluation}

\begin{figure}[H]
    \centering
    \includegraphics[height=0.2\textwidth, width=0.9\textwidth]{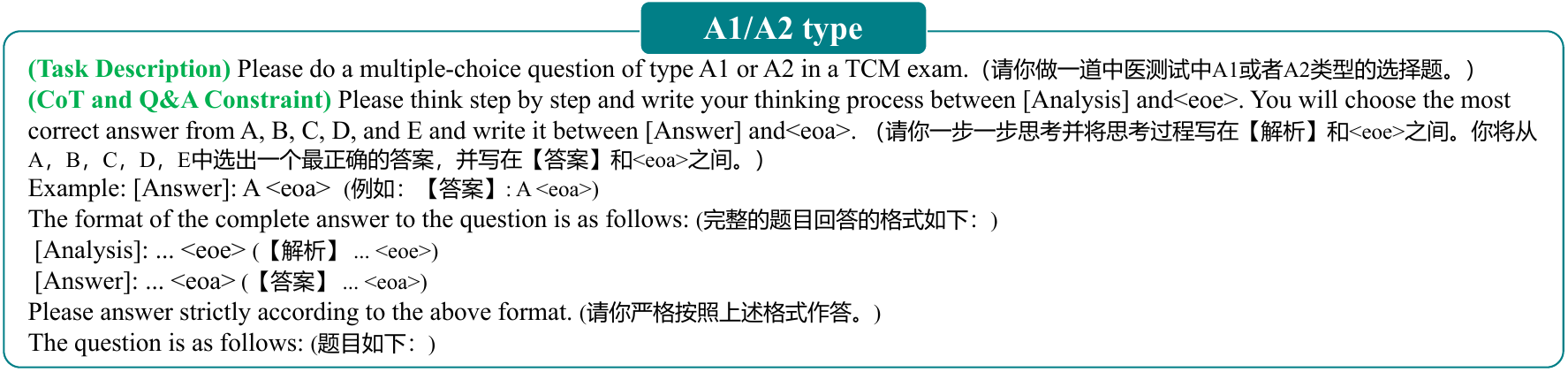}
    \caption{The zero-shot prompt template target output formats for evaluating A1/A2 type of questions.}
    \label{fig:E_A12}
\end{figure}

\begin{figure}[H]
    \centering
    \includegraphics[height=0.33\textwidth, width=0.9\textwidth]{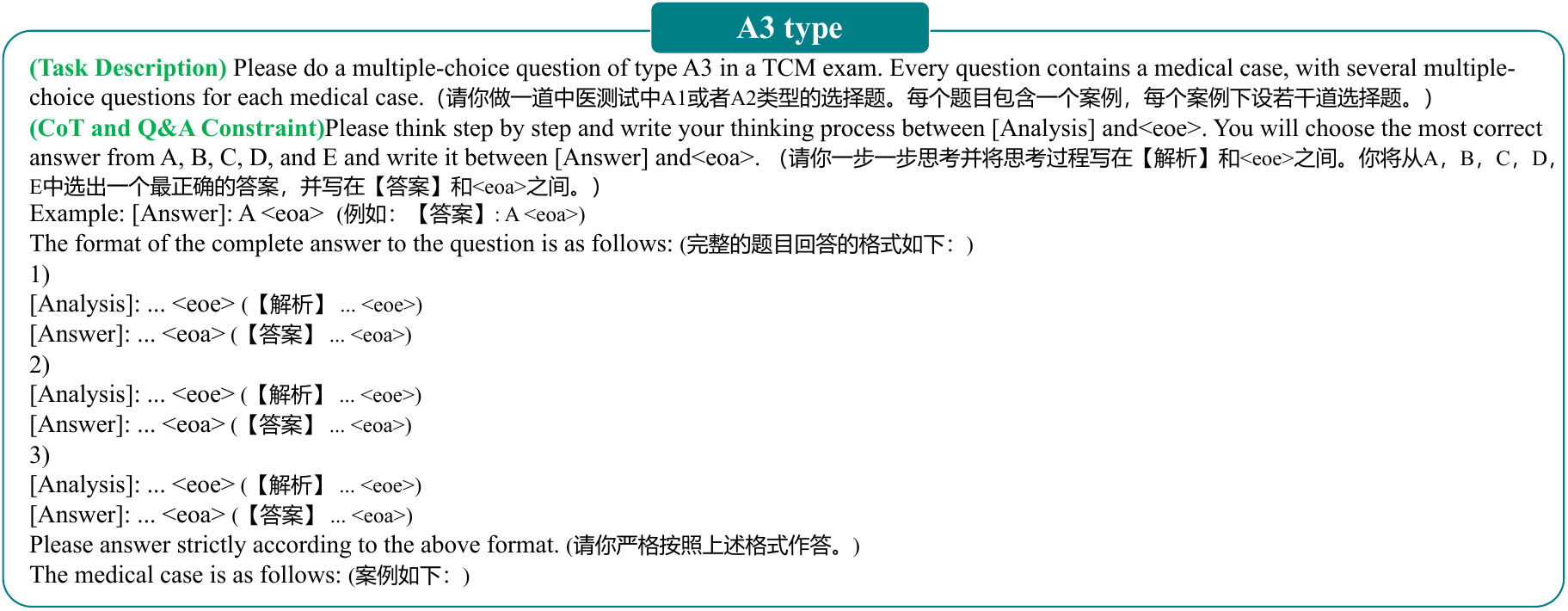}
    \caption{The zero-shot prompt template target output formats for evaluating A3 type of questions.}
    \label{fig:E_A3}
\end{figure}

\begin{figure}[H]
    \centering
    \includegraphics[height=0.25\textwidth, width=0.9\textwidth]{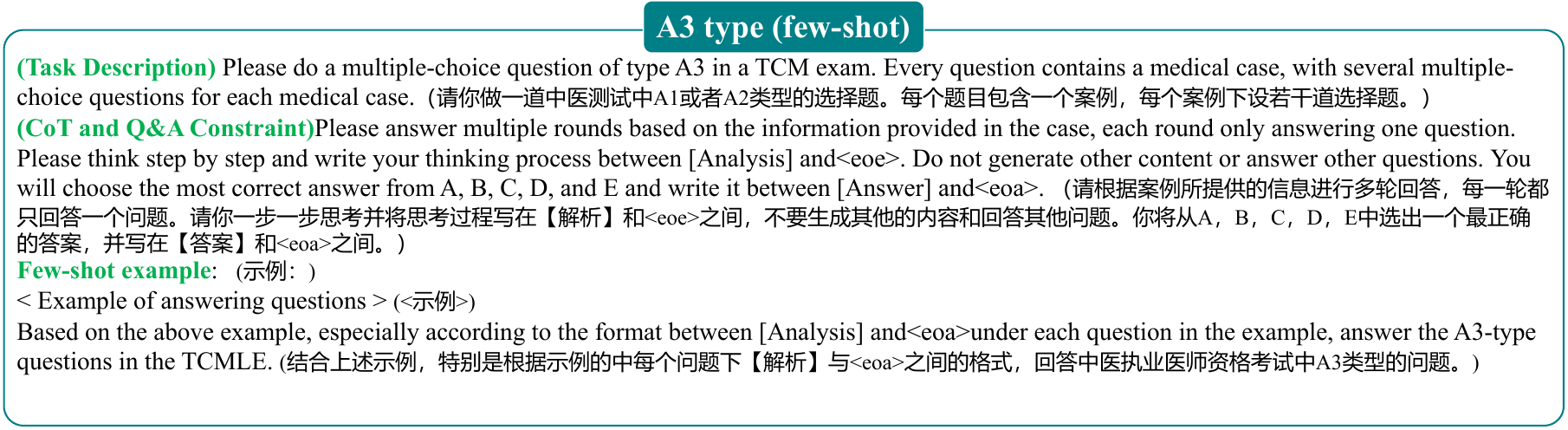}
    \caption{The few-shot prompt template target output formats for evaluating A3 type of questions.}
    \label{fig:E_A3-fs}
\end{figure}

\begin{figure}[H]
    \centering
    \includegraphics[height=0.33\textwidth, width=0.9\textwidth]{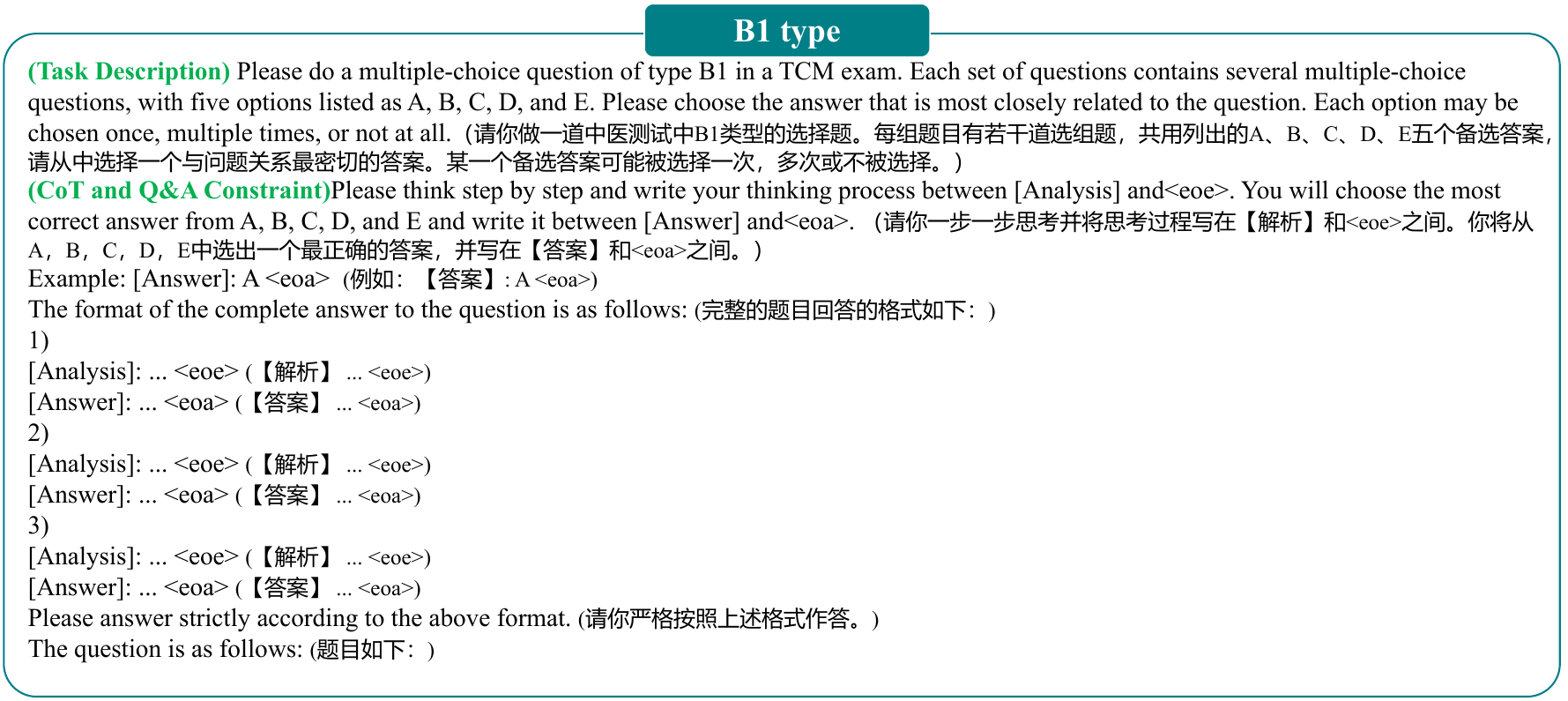}
    \caption{The zero-shot prompt template target output formats for evaluating B1 type of questions.}
    \label{fig:E_B1}
\end{figure}
\end{document}